\documentclass[conference]{IEEEtran}
\IEEEoverridecommandlockouts
\usepackage[belowskip=-10pt,aboveskip=0pt]{caption}
\usepackage{etoolbox}
\usepackage{enumitem}
\usepackage{tabularx} 
\usepackage{cite}
\usepackage{amsmath,amssymb,amsfonts}
\usepackage{url}
\usepackage{amssymb}
\usepackage{pifont}

\usepackage{algorithmic}
\usepackage{graphicx}
\usepackage{textcomp}
\usepackage{xcolor}
\def\BibTeX{{\rm B\kern-.05em{\sc i\kern-.025em b}\kern-.08em
    T\kern-.1667em\lower.7ex\hbox{E}\kern-.125emX}}
\usepackage{amssymb}
\usepackage{xcolor}

\newcommand{\cmark}{\ding{51}}%
\newcommand{\xmark}{\ding{55}}%

\newcommand{\new}[1]{\textcolor{black}{#1}}
\newcommand{\camera}[1]{\textcolor{black}{#1}}

\begin{document}

\title{\textcolor{blue}{CFU Playground: A Full-Stack Framework for Design and Evaluation of TinyML Accelerators}}

\title{CFU Playground: Full-Stack Open-Source Framework For TinyML Acceleration On FPGAs}

\title{CFU Playground: Full-Stack Open-Source Framework For TinyML Acceleration}

\title{\camera{CFU Playground: Full-Stack Open-Source Framework for Tiny Machine Learning (TinyML) Acceleration on FPGAs}}

\author{
\camera{
Shvetank Prakash\textsuperscript{$\ast$}
Tim Callahan\textsuperscript{$\dagger$}
Joseph Bushagour\textsuperscript{$\S$}
Colby Banbury\textsuperscript{$\ast$}}\\
[.2em]
\camera{
Alan V. Green\textsuperscript{$\dagger$} 
Pete Warden\textsuperscript{$\gamma$}
Tim Ansell\textsuperscript{$\dagger$}
Vijay Janapa Reddi\textsuperscript{$\ast$}}\\[.4em]
\camera{
\it 
\textsuperscript{$\ast$}Harvard University
\textsuperscript{$\dagger$}Google 
\textsuperscript{$\S$}Purdue University 
\textsuperscript{$\gamma$}Stanford University 
}
}

\maketitle

\begin{abstract}

Need for the efficient processing of neural networks has given rise to the development of hardware accelerators. The increased adoption of specialized hardware has highlighted the need for more agile design flows for hardware-software co-design and domain-specific optimizations. In this paper, we present CFU Playground--- a full-stack open-source framework that enables rapid and iterative design \new{and evaluation} of machine learning (ML) accelerators for embedded ML systems. \new{Our tool provides a completely open-source end-to-end flow for hardware-software co-design on FPGAs and future systems research.}
This full-stack framework gives the users access to explore \new{experimental} and bespoke architectures that are customized and co-optimized for embedded ML. Our rapid, deploy-profile-optimization feedback loop lets ML hardware and software developers achieve significant returns out of a relatively small investment in customization. Using CFU Playground's design \new{and evaluation} loop, we show substantial speedups between 55$\times$ and 75$\times$. The soft CPU coupled with the accelerator opens up a new, rich design space between the two components that we explore in an automated fashion using Vizier, \camera{an open-source} black-box optimization service.

\end{abstract}

\section{Introduction}

\new{Tiny machine learning (TinyML) is a fast-growing field at the intersection of ML algorithms and low-cost embedded systems. It enables on-device sensor data analytics (vision, audio, IMU, etc.) at ultra-low-power consumption. Processing data close to the sensor allows for an expansive new variety of always-on ML use-cases that preserve bandwidth, latency, and energy while improving responsiveness and maintaining privacy \cite{yang2017method}. Given the need for energy efficiency when running ML on these embedded platforms, custom processor support and hardware accelerators for such systems could present the needed solutions. However, the field of ML is still in its infancy and fast-changing. Thus, it is desirable to avoid a massive non-recurring engineering (NRE) cost upfront, especially for low-cost embedded ML systems. Building ASICs is both costly and time-consuming. Moreover, since embedded systems are often task-specific, there is an opportunity to avoid general-purpose ML accelerators and instead explore task and model-specific ML acceleration methods. This setting presents the need for an agile design space exploration tool that allows us to adapt to the changing landscape of ML and hardware. 
}

\begin{figure}[t]
  \centering
  \includegraphics[width=\linewidth]{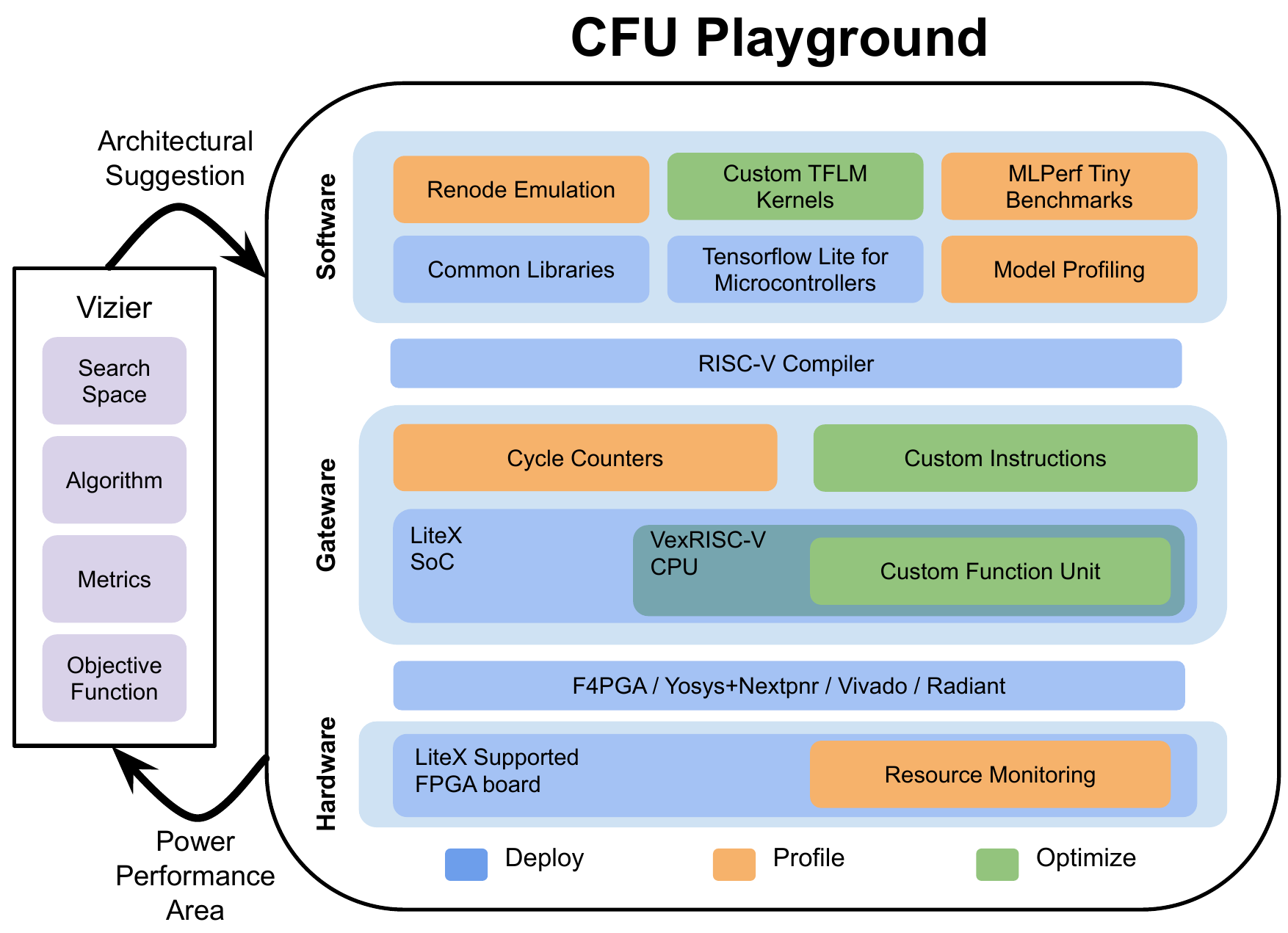}  
  \caption{CFU Playground allows users to design and evaluate model-specific ML enhancements to a ``soft'' CPU core. The Playground is wrapped around \camera{Vizier}, \camera{an open-source} black-box optimization service, to enable ML-driven design space exploration.
  }
  \label{fig:overview} 
\end{figure}

To enable holistic hardware-software co-design and \new{evaluation of} domain-specific performance optimizations easily, we present CFU Playground.\footnote{CFU Playground is available at \camera{\url{www.github.com/google/CFU-Playground}.}}
It is a full-stack open-source framework for iteratively (deploy$\rightarrow
$profile$\rightarrow
$optimize) exploring the design space of lightweight accelerators in an agile manner (Figure~\ref{fig:overview}). 
The framework is unique in that it couples together various open-source software (TensorFlow Lite Micro\camera{/TFLM}, GCC), open-source RTL generation IP and toolkits (LiteX, VexRiscv, Migen, \camera{Amaranth}), and open-source FPGA tools for synthesis, place, and route (yosys, nextpnr, \camera{F4PGA/}SymbiFlow, etc.). 
By using open source for the entire stack, we enable the end-user to \emph{customize and co-optimize hardware and software}, resulting in a specialized solution unencumbered by potential licensing restrictions and not tied to a particular FPGA, board, or vendor. CFU Playground yields large returns out of a relatively small investment in customized hardware and is useful for the long tail of low-volume applications. 

Yet another novelty of CFU Playground is in its \textit{ability to design custom function units (CFUs)} for distinct ML operations. CFUs represent a novel design space that balances acceleration with flexibility and reduces the overhead associated with discrete accelerators. The full-stack solution presented with our hardware-in-the-loop evaluation process works out-of-the-box. It also accounts for end-to-end bottlenecks that may arise elsewhere in the stack (software overheads, pre-processing, etc.) but are often ignored when designing in isolation. From an initial working, non-customized solution, the user can incrementally specialize individual (hardware or software) components to improve their application's performance. Due to the lightweight nature of CFUs, one can develop and make changes quickly as compilation and deployment targeting embedded ML platforms is rapid.

We use the framework to demonstrate how to design CFUs, extending an FPGA-based RISC-V core in a way that is fully integrated with the rest of the system software stack. The primary reason CFUs are suitable for ML inference is that there are often a few small yet critical hotspots. A small amount of custom hardware that exploits the bit-level flexibility of an FPGA can help accelerate large portions of execution time. A tightly integrated CFU allows us to leave complexity, setup, and outer loops in the software while efficiently tackling the core computational bottlenecks in the datapath. Moreover, as we demonstrate, CFUs allow us to incrementally grow the unit until it almost becomes a full-blown ML accelerator. Using our agile design flow, we show how to accelerate the convolution operation of MobileNetV2 via a combination of optimizations such as loop unrolling, SIMD multiply-accumulate, and pipelining to a 55$\times$ speedup. \new{It took a senior engineer only working part-time on the project about five weeks to achieve this massive speedup.}

Additionally, with a Keyword Spotting application, we show how, due to the nature of the RISC-V soft CPU, the core can not only be extended (new instructions added through the use of a CFU) but also tailored (e.g., cache sizes modified) to meet the platform’s constraints given limited resources. This also enables the user to perform a design space exploration between resources allocated to the CFU, CPU, or memory system. Exploitation of this ability along with strategic use of parallelism and pipelining led to an overall speedup of 75$\times$. 
\new{It took an undergraduate-level intern with minimal FPGA and hardware experience under four weeks to achieve this speedup, owing to the various abstraction layers. }

In summary, our contributions are as follows:
\begin{itemize}[topsep=1pt, itemsep=1pt]
    \item An out-of-the-box, \textit{full-stack framework} that fully integrates open-source tools across the entire stack to facilitate rich community-driven ecosystem \new{research} and development. 
    \item An agile methodology to iteratively design \new{and evaluate} \textit{tightly-coupled, bespoke accelerators} for resource-constrained, latency-bound TinyML applications.
    \item We 
    demonstrate novel \textit{model-specific resource allocation trade-offs} between the CFU, CPU, and memory system that enable optimal ML performance on resource-constrained FPGA platforms for two important use cases.
    \newline 
    \item We bundle CFU Playground with Vizier, an open-source \textit{black-box optimizer} from Google to enable automated design space exploration of the CPU paired with a CFU.
    
\end{itemize}


\renewcommand\thesubsubsection{\Alph{subsection}}

\section{\new{Background \& Motivation}}

\subsection{Tiny Machine Learning (TinyML)}

\new{TinyML is the deployment of machine learning (ML) algorithms onto low-cost, low-power, and resource-constrained microcontroller (MCU) systems. TinyML enables on-device ML and achieves this using a fraction of the compute resources needed for traditional ML systems. Table~\ref{table:MLSystemsComparison} compares TinyML with traditional BigML (such as cloud and mobile systems) and shows how TinyML is orders of magnitude smaller in terms of compute, memory, storage, power, and cost. The ML models running on-device can be used for intelligent, on-device sensor analytics, unlocking always-on ML use-cases. While there are many benefits to TinyML, the heterogeneity of MCU hardware and limited resources available on them presents new challenges.} 

\new{MCUs typicallly only have ten to a few hundred KBs of SRAM and one to two MBs of Flash. This severely limits the size of the on-device ML models. Compute, memory, and storage are also often tightly integrated, limiting the ability to adapt the resources flexibly to various needs. While these systems have proved capable of basic ML tasks such as Keyword Spotting, Visual Wake Words, and Anomaly Detection~\cite{banbury2021mlperf}, specialized hardware is needed to support more advanced applications while maintaining a low-power operating point.}




\renewcommand{\arraystretch}{1.5}
\begin{table}[t]
\resizebox{\columnwidth}{!}{
    \begin{tabular}{|c|c|c|c|c|c|}
     \hline
        \shortstack{Platform} &
        \shortstack{Freq.} & 
        \shortstack{Memory} &
        \shortstack{Storage} & 
        \shortstack{Power} & 
        \shortstack{Price} \\
     \hline
     \hline
     \hline
        Cloud & 
        GHz & 
        10+GB & 
        TBs-PBs & 
        100 W-kW & 
        \$1000+  \\ 
     \hline
        Mobile & 
        GHz & 
        Few GB & 
        GBs & 
        1$\sim$10 W & 
         \$100+ \\ 
     \hline
         \bf Tiny & 
         \bf MHz & 
         \bf KBs & 
         \bf Few MB &
         \bf $\boldsymbol\mu$W $\sim$ mW &
         \bf $<\$$10 \\
     \hline
    \end{tabular}

}
\vspace{5pt}
\caption{Cloud \& Mobile ML Systems vs. TinyML systems.}
\label{table:MLSystemsComparison}
\end{table}

\subsection{Need for Agile and Full-Stack \new{Research} Frameworks}

\new{TinyML presents numerous challenges that we believe an open-source ecosystem can address through systems research and development}. 
ML acceleration on microcontroller-class hardware is a new area, so TinyML deployment tools and runtime environments are often vendor-specific, leading to software fragmentation across the hardware platforms~\cite{banbury2020benchmarking}.  


\new{Moreover, many of the TinyML tools are not publicly accessible. This lack of access makes isolating and comparing accelerator performance challenging. In addition, hardware customization is an iterative process, especially for ML development in which the algorithm and model itself typically undergo refinement. For example, there is often a back-and-forth between the ML team and processor implementation team--``What if we quantize this layer to 4-bits?''--to achieve a solution that meets both the performance goals as well as the embedded system's resource constraints. }


\new{To that end, there is a clear and present need for agile design space exploration tools, despite prior work (which we discuss extensively in Section~\ref{sec:related}). While simulators are  useful and good for functional correctness, getting accurate performance estimates is difficult. Modeling all features of a system (e.g., multiple clocks, asynchronous interfaces, mixes of memories with different bandwidth and latency, etc.) accurately can result in a slow simulation also. When running with \textit{hardware-in-the-loop}, users know that measurements collected are not overlooking a key performance factor.}

\new{Additionally, if one is successful speeding up computation, the system often becomes memory or I/O limited. Therefore, modeling and evaluating the ML accelerator's performance from a full-stack system perspective is also important. Modeling bus latencies including contention and arbitration is inexact at best using simulators. Full-system cycle-accurate simulations are too slow for agile design.}


\subsection{Design Space Exploration on FPGAs}


\new{FPGAs open up complex design space exploration for customized microarchitectures that extend beyond the ML accelerator and include the CPU. 
The opportunity exists with an FPGA platform to customize the processor to adapt it to perform the application's computation efficiently. ML computations tend to be regular and repetitive, the same sequence of operations repeated millions of times. We can focus on improving the execution of this ``hot” computation while using standard instructions to perform the rest. A small amount of custom hardware for these hotspots that exploit the bit-level flexibility of FPGAs can translate to large improvements.}


\new{An FPGA platform also allows for an accelerator unit to be tightly coupled into the CPU pipeline that can be easily invoked by adding new custom instructions that complement the CPU’s standard functions. It is an alternative to treating ML accelerators as discrete processing units, which suffer from ``AI Tax''~\cite{buch2021ai,richins2021ai}. 
Along with the ease of programming, tight coupling is beneficial for latency-bound applications~\cite{hanawa2013tightly} as needed for TinyML~\cite{banbury2021mlperf}.   
Furthermore, the bit-level flexibility of FPGAs allows efficient implementation of operations on small data sizes and non-standard data representations, and even packing, unpacking, and converting between data types.} 


%
\new{An FPGA-based platform has benefits other than computational efficiency. In embedded ML applications, security is often paramount due to their always-on nature.  A soft CPU provides transparency so that one knows exactly what is inside, in contrast to needing to place blind trust in the supply chain of the chips acquired. Another protection against supply chain uncertainty is that if the original FPGA target becomes unavailable, with minimal work you can switch to another device or vendor since your ``processor,” including CPU and custom function unit(s), is just Verilog that can be deployed on any FPGA with enough resources.}

\section{CFU Playground Overview}


There is a need for accessible and agile tools such as CFU Playground \new{for experimental TinyML and systems research}, which combine a collection of gateware, software and hardware pieces, as shown in  Figure~\ref{fig:overview}. In this section, we describe \new{the design of our framework and} each of these major components in greater detail.

\begin{figure}[t]
\centering
\begin{minipage}{.5\columnwidth}
  \centering
  \includegraphics[height=1.25in]{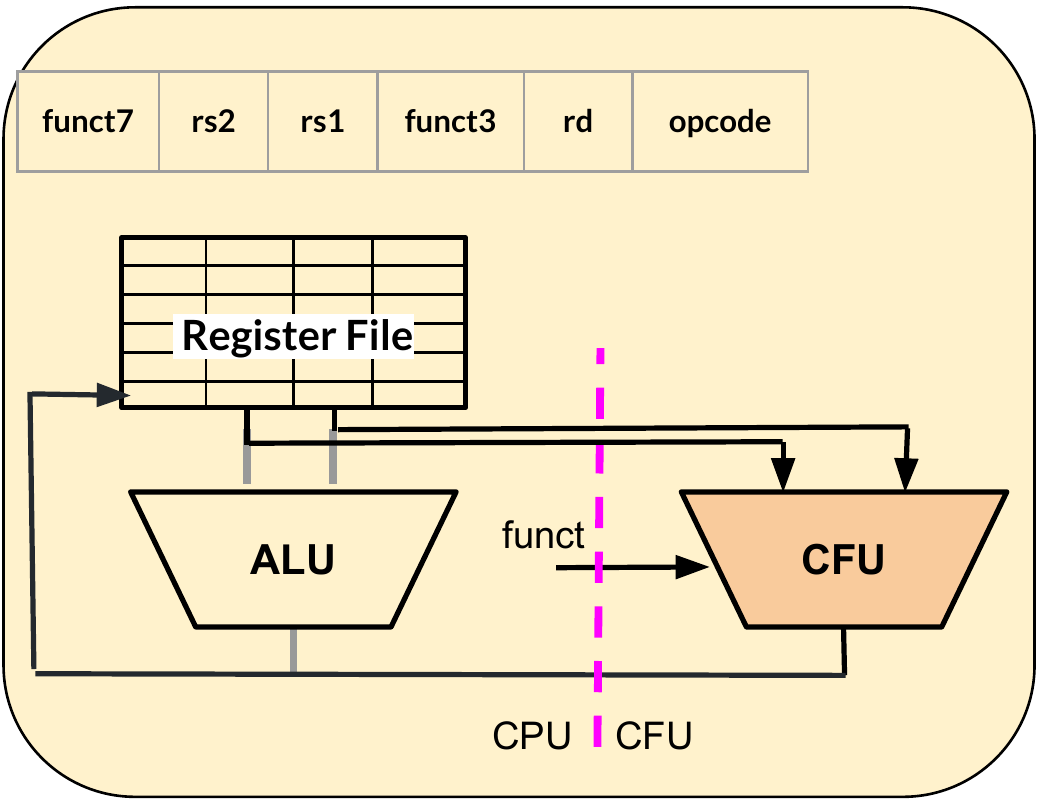}
  \caption{Custom Func. Unit.}
  \label{fig:cfu} 
\end{minipage}%
\begin{minipage}{.5\columnwidth}
  \centering
  \includegraphics[trim=0 0 0 10, clip, height=1.25in]{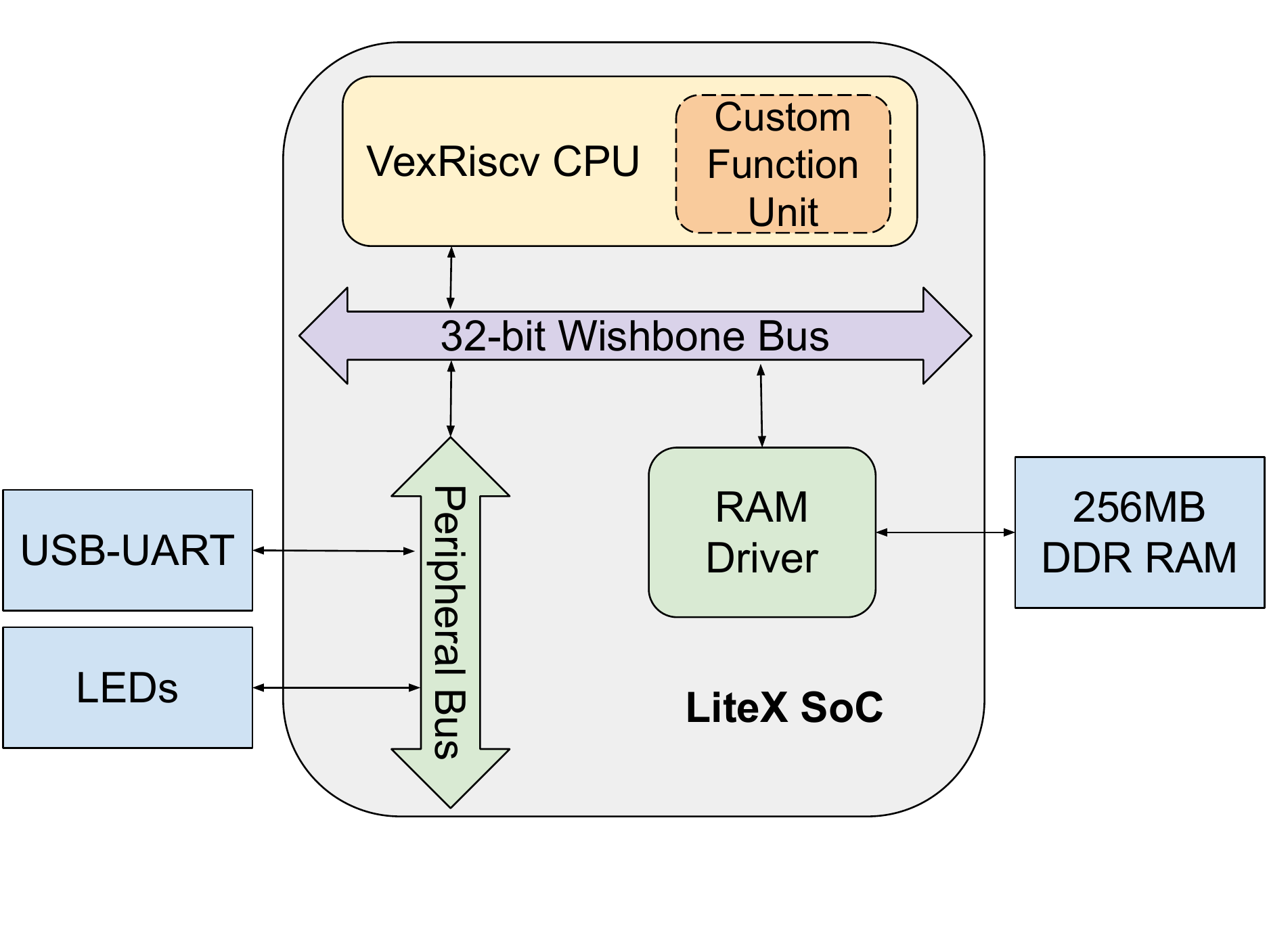}
  \caption{LiteX SoC.}
\label{fig:litex-soc} 
\end{minipage}
\end{figure}

\subsection{Custom Function Unit (CFU)}
The CFU\footnote{\new{CFUs are part of the Composable Custom Extensions Project~\cite{cfu}. The project's proposal aims to create a solution the allows for creation of reusable extensions for any hardware without having to wait years for ratification. Thus, the CFU approach combines the advantages of standard and custom extensions.}} is a small piece of custom logic added in hardware to extend the CPU's datapath to accelerate a discrete function determined by the developer. A custom function unit (CFU) can improve program hotspot execution in many ways. A CFU can specialize operations for constant operands, perform multiple operations per cycle through parallel computation of independent operations, perform a fusion of successive operations within a cycle, or pipeline the computations across multiple cycles. Flexible, configurable storage also allows the data to be stored and reused locally, thus reducing unnecessary data movement. The user can also add and perform a combination of these {iteratively} to progressively improve performance.

The CFU follows the RISC-V R-format in which it receives two operands from the register file and writes one result back. A CFU can support state, multiple custom instructions, and pipelining. Figure \ref{fig:cfu} depicts the architecture of the CFU in relation to the rest of the CPU.
The boundary between the CPU and CFU is strictly logical. The current implementation flattens the design and optimizes, places, and routes it all together.




\subsection{Gateware}

CFU Playground incorporates a CFU into a System-on-Chip (SoC) on an FPGA to capture the full-stack system effects of accelerating ML models. See Figure \ref{fig:litex-soc}.
Its gateware is built upon the LiteX framework~\cite{litex}. 
LiteX provides a convenient and efficient infrastructure to create FPGA soft cores and SoCs. 
For any board to be used in CFU Playground, 
it must first have a LiteX description.
Most popular commercially-available boards already have a 
description in the crowd-sourced LiteX boards library.
For a private prototype board, the description can be created and maintained locally.   

\camera{The CFU Playground user has access to the full set of LiteX options for customizing the SoC.  For example, on boards that support an L2 cache, the L2 cache size can be changed simply by adding this line
to the project Makefile:}

{\tt \camera{export EXTRA\_LITEX\_ARGS='--l2-size=16384'}}

\pagebreak
\camera{To make that change for all projects when using that board, add this to the board's customization in one line of Python:}

{\tt \camera{args.l2\_size = 16384}}

The soft core used in CFU Playground is VexRiscv, an \camera{open-source} implementation of a RISC-V CPU in SpinalHDL~\cite{vexriscv}. \camera{VexRiscv was the winner of the soft CPU contest at the RISC-V Summit and has been optimized for FPGAs.} The design of VexRiscv is highly configurable, providing the ability to easily plugin or remove many different features for performance and functionality such as pipelining stages, caches, and floating point units. This customization ability lends itself well to enabling the design space exploration of CPU vs. CFU. Currently VexRiscv is the only soft CPU that supports the specific CFU interface used. Future work aims to add the CFU interface to other \camera{RISC-V} 
CPUs. Nonetheless, this still gives the developer a wide array of VexRiscv designs to explore.

\subsection{Hardware}

The gateware for CFU Playground is adaptable to a wide range of FPGAs. \camera{Table~\ref{table:FPGABoards} lists a few of the supported FPGA boards.}
The gateware can fit on a board as small as Fomu~\cite{fomu} for TinyML prototyping, which is 1 $ cm^2 $
and fits in a USB port.
However, when more resources are available, a more powerful soft CPU and CFU can be built,
and with more memory, larger models can be run. Thus, the system is inherently scalable.

The minimum requirements for the board and its FPGA include the following. There must be some means of creating a TTY / UART connection to interact with software on the board. The FPGA needs enough resources to build VexRiscv CPU variants. The system must have enough RAM (on the FPGA or externally on the board) to provide working memory for the software. There must be sufficient RAM or ROM to hold code and constant data like the TensorFlow Lite model.

\new{It is difficult to put an exact set of minimum requirements on any of the resources for the evaluation hardware 
since they are interchangeable to some degree. For example, the CPU can be built 
purely out of LUTs, but it will require many fewer LUTs if it can use a block RAM for its register file and a DSP block for its multiplier.}

CFU Playground currently supports the Xilinx 7-Series as well as the Lattice iCE40, ECP5, and CrossLink FPGAs. The Fomu with the iCE40UP5k FPGA is close to the smallest usable board. It features
5280 logic cells, 128kB on-chip large RAM, 30 512-byte block RAMs, 8 16-bit x 16-bit DSP/multiplier blocks, and fits into a USB slot. Verilog for the CFU is compiled to a bitstream by either an open-source (e.g., F4PGA/SymbiFlow~\cite{symbiflow}) or vendor-supplied FPGA toolchain. \camera{We note that the CFU, though, can be written in any HDL able to generate Verilog that the user chooses (e.g., Chisel, Bluespec, Amaranth, etc.).}


\begin{table}[t]
\resizebox{\columnwidth}{!}{
\begin{tabular}{|c|c|c|c|c|c|}
\hline
Board &  Look Up Tables (LUTs) & DSPs & RAM & ROM & Sys Clk Freq  \\ \hline
Arty A7-100T & 101,440 & 240 & 256MB & 16MB & 100MHz  \\ \hline
Arty A7-35T  & 33,280 & 90 & 256MB & 16MB & 100MHz  \\ \hline
OrangeCrab   & 24,000 & 28 & 128MB  & 16MB &  75MHz \\ \hline
ULX3S (12F)  &  12,000  &   12   &  32MB & 4MB  & 50MHz  \\ \hline
iCEBreaker   & 5280 & 8 & 128kB & 16MB & 24MHz  \\ \hline
Fomu         & 5280 & 8 & 128kB & 2MB & 12MHz  \\ \hline
\end{tabular}
}
\vspace{10pt}
\caption{\camera{Examples of embedded FPGA boards tested and supoorted by CFU Playground. }}
\label{table:FPGABoards}
\end{table}
\subsection{Software}

To invoke the CFU hardware, custom instructions must be added to the CPU’s instruction set. However, it is not the compiler’s responsibility to find uses for the instructions. It only needs to generate them when requested by the user. 
\camera{This is same process used in each of the hardware optimized kernels in TensorFlow Lite for Microcontrollers (e.g. CMSIS-NN~\cite{lai2018cmsis}) and allows us to use existing RISC-V compiler support, which we value for rich open-source ecosystem research and development in the future.} Therefore, we can 
\camera{take} 
a stock RISC-V GCC toolchain, coupled with a macro we provide that expands to ``asm'' directives to generate the encoded custom instructions where necessary. The macro takes 4 arguments, and returns one result:

\begin{center}
\begin{small}
\texttt{q = cfu\_op(funct7, funct3, a, b);}
\end{small}
\end{center}

The macro directly generates the encoded 32-bit value, so not even the assembler needs modification. 
``\texttt{funct7}" and ``\texttt{funct3}" are 7-bit and 3-bit fields respectively that specify the opcode of the custom instruction. They must be compile-time constant expressions.  ``\texttt{a}" and ``\texttt{b}" are the C/C++ 32-bit integer variables used as operands for the instruction, and a 32-bit result is returned. The macro gets information from the compiler regarding the register locations of operands and result (\texttt{a}, \texttt{b}, and \texttt{q} in this example) to generate the correct bit fields for register sources and destination in the encoded instruction, so extra register-to-register copies are not needed. For readability, the user may define a macro for each custom instruction, in terms of the base \texttt{cfu\_op( )} macro as shown:

\begin{small}
\noindent \new{\texttt{\#define single\_popcount(a) cfu\_op(1,1,(a),0)}}
\newline
\new{\texttt{\#define bit\_reverse(a)\ \ \ \ \ cfu\_op(1,2,(a),0)}}
\newline
\texttt{\#define simd\_add(a, b)\ \ \ \ \ cfu\_op(1,3,(a),(b))}
\end{small}

\new{As with any inline assembly instructions, these custom instruction macros can be intermingled with regular C code.}

\new{When disassembling, the toolchain knows nothing about the custom instructions, not even which bits specify registers, so it simply writes out the 32-bits as a hexadecimal value. CFU Playground converts the hexadecimal to readable assembly including the source and destination registers. }

\new{This is an example of custom instruction macros intermixed with regular C code:}
\texttt{
~\newline
~\newline
  int *x;  \newline
  int b;   \newline
  ...      \newline
  t1 = *x; \newline
  t2 = cfu\_op(0, 0, t1, b); \newline
  t3 = cfu\_op(1, 0, t2, b); \newline
  *x = t3; \newline
}

\new{And the following is the result of the example compiled and then disassembled, illustrating how the custom instructions can be used with no extra overhead:}
\texttt{
~\newline
~\newline
400001a0:       00812783            lw        a5,8(sp) \newline
400001a4:       00d7878b            cfu[0,0]  a5, a5, a3 \newline
400001a8:       00d7978b            cfu[1,0]  a5, a5, a3 \newline
400001ac:       00f12423            sw        a5,8(sp) \newline
}

It is the user's responsibility to call the custom operations in their code. The custom instruction macros can be intermixed with regular C code, similar to any other C/C++ operation. TensorFlow Lite for Microcontrollers (TFLite Micro\camera{/TFLM})~\cite{david2020tensorflow} is the inference framework that CFU Playground uses for the deployment of the neural network. The user must provide an optimized kernel that uses the new custom instructions to realize the runtime performance improvements. 
\new{Since CFU Playground and TFLite Micro use a source-file overlay mechanism, no modifications are necessary to the inference framework. Simply implementing the optimized kernel in a separate file will cause the build to pick up the new implementation. Typically, just one or two TFLite operation types are targeted for acceleration, so only the kernels that implement those op types need to be modified and optimized. TFLite Micro uses an interpreter to perform inference and can call the specifically optimized kernels for the individual network layers. }

\new{Moreover, TFLite Micro provides flexible deployment at the cost of a reasonably minimal memory and latency overhead compared to vendor-specific frameworks. Thus, if the user created new instructions to speed up the \texttt{CONV\_2D} operation type, they would start by copying the TFLite library file to the location under the current project directory: 
\camera{\fontsize{6.2}{12}\texttt{ ./src/tensorflow/lite/kernels/internal/reference/integer\_ops/conv.h}}
\newline
And then they would modify it to use the custom instructions.}

\new{The TFLite library kernels are general in that they can handle any legal parameterization of the TF operations, and any legal input or output tensor shapes. However, the user may choose to handle only particular parameterizations with their modified, optimized kernel. In this case, the modified version of the kernel usually contains a check at entry; if the parameters fall outside of the supported set, then execution is delegated to the general kernel. At deployment, though, the general version can be removed if the model(s) contain only parameterizations that are supported by the specialized kernel.}

\subsection{Testing, Simulation, and Benchmarking}

The menu-driven software contains kernel-level unit tests from the TFLite Micro library. It also contains full-inference golden tests, with set inputs and expected outputs for each provided model. 
Additional models can be added as desired, which we demonstrate in the evaluation section (Section~\ref{sec:eval}).

To support debugging and testing, users can write a software emulation of their CFU, using the high-level C programming language, that is functionally equivalent but of course much slower, which can be swapped in for the real CFU. The software emulation function has multiple uses. For example, random or directed CFU-level unit tests running on the FPGA board can feed the same sequence of inputs to both the real CFU and to the software emulation, and expect to see the same sequence of outputs to test for correctness. When execution using the real CFU appears to be incorrect, the CFU can be replaced by the software emulation to determine if the problem is in the code that uses the custom instructions or in the implementation of the user-defined CFU. Moreover, the application running on the FPGA board has access to \texttt{printf()} and terminal output, so printing variable values during execution can be done as an alternative debugging method. Some boards support a debug bridge to connect a debugger to the running system. In this case, a debug-enabled VexRiscv variant must be used and such variants readily exist.

\camera{ 
CFU Playground also contains simple tests for running the actual custom instructions against the software emulation with random, exhaustive, or directed inputs.   These are written in C and run on the board's SoC.   Directed tests are more appropriate for complex, stateful CFUs, and the existing code can be easily adapted as appropriate for each project.
}

\camera{For complex CFUs written in \camera{Amaranth HDL (previously known as nMigen~\cite{nmigen})}, an open-source Python-based toolkit for RTL design, 
unit test facilities are provided in Python itself. Given Amaranth's unit testing, and the other testing methodologies described below, Verilog-level, simulation unit testbenches for the CFU may not be needed, but the user can certainly create their own.}

CFU Playground also supports the Renode emulator \cite{renode}, which emulates the physical hardware system. While following a hardware-in-the-loop process and running on a physical board is recommended by CFU Playground, a Renode emulation can be used to test CFUs on other boards without having them. Renode performs ISA simulation of the CPU, combined with cycle-accurate Verilog simulation of the CFU, \camera{thus testing functional correctness of the CFU implementation}. It also simulates the RAM, ROM, and UART. The Renode emulator also allows us to capture the waveforms from the CFU operation, which is extremely useful for tracking down errors in the hardware design of the user-defined CFU.

\new{Moreover, to help the field progress, there has been an effort to benchmark these heavily resource-constrained TinyML systems to drive standardization and innovation in the industry~\cite{banbury2020benchmarking}. CFU Playground comes packaged with the MLPerf Tiny~\cite{banbury2021mlperf} deep learning workloads for benchmarking purposes.}

\new{There are two interesting use cases in which an agile design flow like ours can help support this benchmarking ecosystem and research community. The first occurs when a user has a specific idea for how to accelerate an existing model. Being able to rapidly prototype and benchmark that idea before going through the expensive and time consuming process of building an ASIC accelerator can enable exploration of new \textit{hardware} architectures. The second use case addresses the hardware lottery problem~\cite{hardwarelotteryproblem} and can enable exploration of new \textit{model} architectures. One of the reasons CNNs exploded was because of their convenient mapping to existing hardware (i.e., GPUs). However, if a user has an idea for a novel model architecture for TinyML that is not well-suited for existing hardware, CFU Playground could be used to quickly bootstrap hardware for the new model architecture so that it could be fairly benchmarked against existing model architectures that already have mature hardware support. This could assist in proving their model is efficient given the proper hardware.}

\subsection{Design Space Exploration}
\label{sec:dse}

We package CFU Playground with the open-source version of Vizier~\cite{oss_vizier}, a black-box optimization service based on the Google's internal Vizier framework. Using this service, our framework enables automated design space exploration (DSE) of the CPU coupled with CFU. The DSE parameters could include branch predictors types (static, dynamic, dynamic target), custom function units (SIMD, MAC, etc.), I- and D-cache sizes, multipliers, dividers, shifters, etc. \camera{The user can readily make these parameters  available in Python to Vizier's search space using the API:} 

\camera{\texttt{problem.search\_space.select\_root()
\indent\indent.add\_categorical\_param(
\newline\indent\indent\indent name='prediction', \newline\indent\indent\indent feasible\_values=['none', 'static', 
\newline\indent\indent\indent'dynamic', 'dynamic\_target']
\newline\indent\indent)}}

\camera{Vizier's service then returns} different configurations to explore based on what the user would like to optimize (e.g., resources or latency). Yosys~\cite{yosysnextpnr}, the open-source synthesis tool used in CFU Playground, calculates the FPGA logic cell resources required by a CPU-CFU design, which are then passed to Vizier. Verilator~\cite{snyder2013verilator}, a cycle-accurate simulator also packaged with the framework, is used to determine the workload latency for Vizier when running experiments at scale in the cloud. The latency can also be measured directly on the board with our hardware-in-the-loop capability. CFU Playground's DSE feature allows users to tune their CPU for their workload and analyze the trade-offs between CFU and CPU components, which is extremely valuable in resource constrained environments. 
Vizier's systematic search is critical for exploring the large and diverse design space, enabled by CFU Playground, in a tractable amount of time. 
\camera{We provide a \texttt{vizier\_dse.py} script that can be executed out-of-the-box to run this design space exploration loop (Figure~\ref{fig:overview}).}

\section{\camera{Evaluation}}
\label{sec:eval}

\camera{In this section, we first present the design and evaluation methodology used in CFU Playground (Section~\ref{sec:dev_life_cycle}). After outlining how we facilitate guided optimizations during experimental design, we present two case studies to evaluate the framework's usability (Section~\ref{sec:case_studies}). Finally, using the accelerator designs from the case studies, we assess the trade-offs between CPU and CFU using automation and highlight the rich design space enabled by CFU Playground (Section~\ref{sec:dseengine}).}

\subsection{\camera{Development Life Cycle}}
\label{sec:dev_life_cycle}

\begin{figure}
    \centering
    \includegraphics[width=\linewidth]{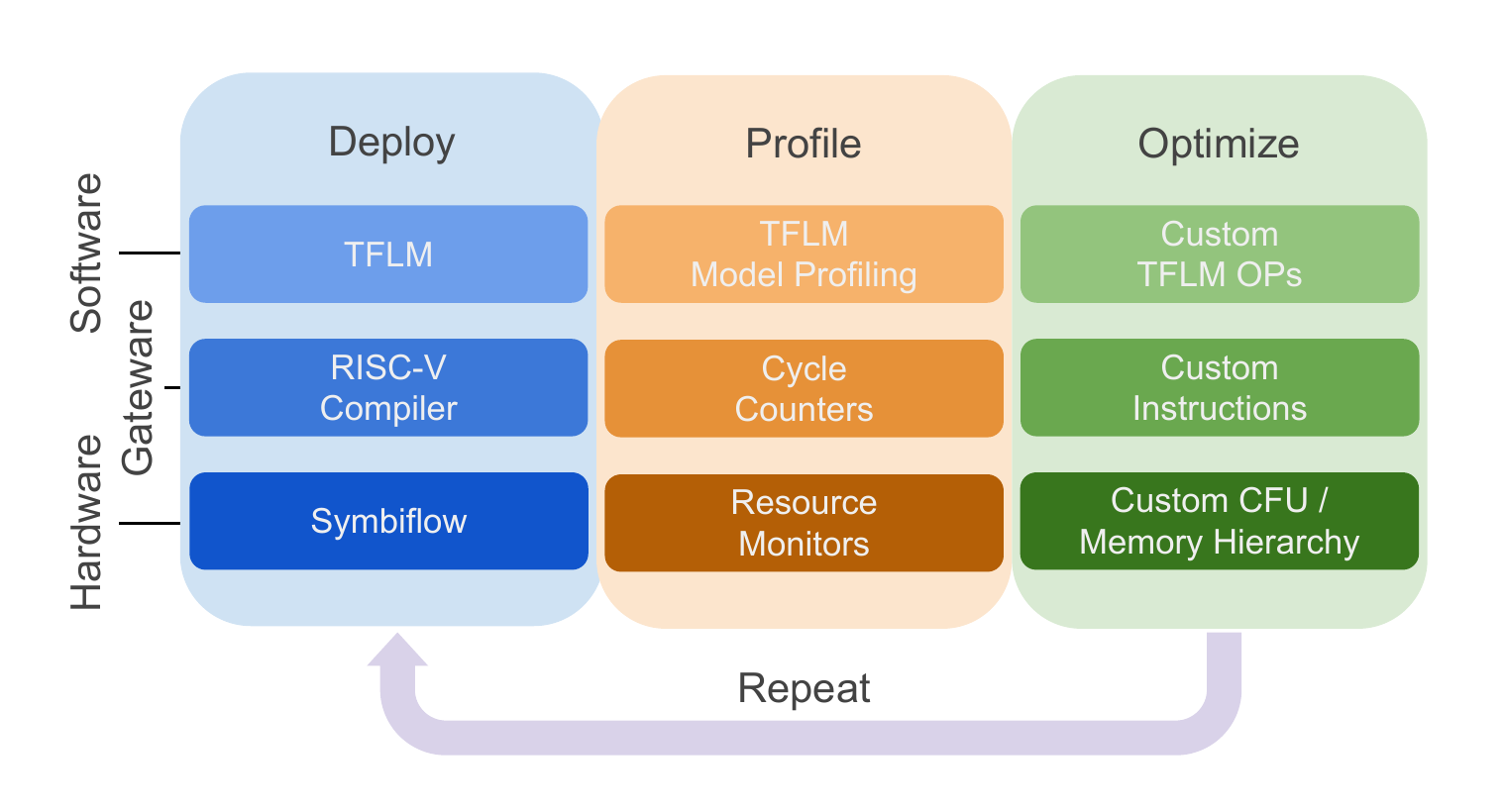}
    \caption{\new{The workflow of CFU Playground that enables systematic development and evaluation of experimental and bespoke cross-stack optimizations.}}
    \label{fig:cfu-design-flow}
\end{figure}

\new{CFU Playground enables a deploy-profile-optimize design and evaluation loop that enable iterative, guided optimization of resource-constrained ML systems.
The toolchain is constructed such that a developer can, at any layer in the stack, easily run their design, measure it's performance at a fine granularity, implement custom optimizations, and repeat.}

\new{Figure \ref{fig:cfu-design-flow} illustrates this evaluation workflow and highlights components of CFU Playground that fulfill a role in the design cycle at each layer in the deployment stack. The full-stack spans across the Software, Gateware, and Hardware components as we have illustrated in Figure~\ref{fig:overview}.}



\textbf{\camera{Software Optimization:}}
\new{CFU Playground enables the developer to profile the model to understand baseline performance. For example, TFLM’s built-in profiling shows the name and running time of each TFLM operation (i.e. \texttt{CONV\_2D}, \texttt{DEPTHWISE\_CONV}, etc.) as it executes.
Cycle counters can then be used to profile parts of the C code for individual operations to identify hot spots.}

\new{Based on the profiling information, the developer can create a custom TFLM op that does things like caching frequently used variables and unrolling loops to get acceleration fairly quickly this way via software optimization alone. }


\textbf{\camera{Hardware Acceleration:}}
\new{After simple software optimizations, the developer can identify opportunities for CFU acceleration with resource monitors that can identify hardware bottlenecks. Examples of such optimizations could include moving loop-invariant variables into the CFU, making SIMD versions of instructions, and making their corresponding custom instructions. Such optimizations can provide further significant speedups in combination with software optimizations.}


\camera{\textbf{Automated Co-design:} After the developer performs software and hardware optimizations, there is still opportunity for more specialization by exploring the design space of CPU + CFU configurations. This allows for co-design between the configurable processor and accelerator to help tailor the user's final design even further towards its specific task. To help assess resource-performance trade-offs between the soft CPU + CFU configurations, the developer can use the framework's design space exploration infrastructure to automatically tune their embedded ML system to meet constraints.}





\subsection{\camera{Case Studies}}
\label{sec:case_studies}

To show how CFU Playground is useful in practice for end-users \new{and experimental designs}, we show the iterative development methodology using two common TinyML use cases: Image Classification (IC) and Keyword Spotting (KWS). 

The IC example showcases how CFU Playground enables iterative hardware-software improvements with ease. The KWS example shows how we can co-optimize the CPU and the CFU together in severely resource-constrained environments. 

We focus on performance optimizations but the methods can also extend to energy-related optimizations. All of these examples and infrastructure are publicly available. \camera{The two baselines mentioned in the following sections were obtained running on a variant of the basic VexRiscv configuration (i.e., with no CFU acceleration).}

\subsubsection{Image Classification Acceleration on Arty}
\label{sec:ic_on_arty}

We present how the CFU Playground makes it easy to deploy$\rightarrow$profile$\rightarrow$optimize and  achieve a speedup of 55$\times$ for the most time-consuming TFLite operator, bringing operator time down from 5.5 seconds to 0.10 seconds per inference. We begin with software optimizations and then move down to hardware, showing the co-design easily enabled by our tool.

%
%

\textbf{Target Objective:} 
The goal was to speed up the operation type that dominated the runtime so that it became insignificant, to show the use of our iterative methodology, which could then be applied to all the other operations in theory as well. 

\textbf{Deploy:} 
We targeted the MobileNetV2 (MNV2) model for efficient image classification and optimized its performance on an Arty A7-35T board, which has a Xilinx XC7A35T FPGA with 256~MB of external DDR3 memory. The model was quantized down to 8-bit integers (int8) as this is what the TFLite Micro inference framework that we used supports.

\begin{figure}[t]
  \centering
  \includegraphics[width=\linewidth]{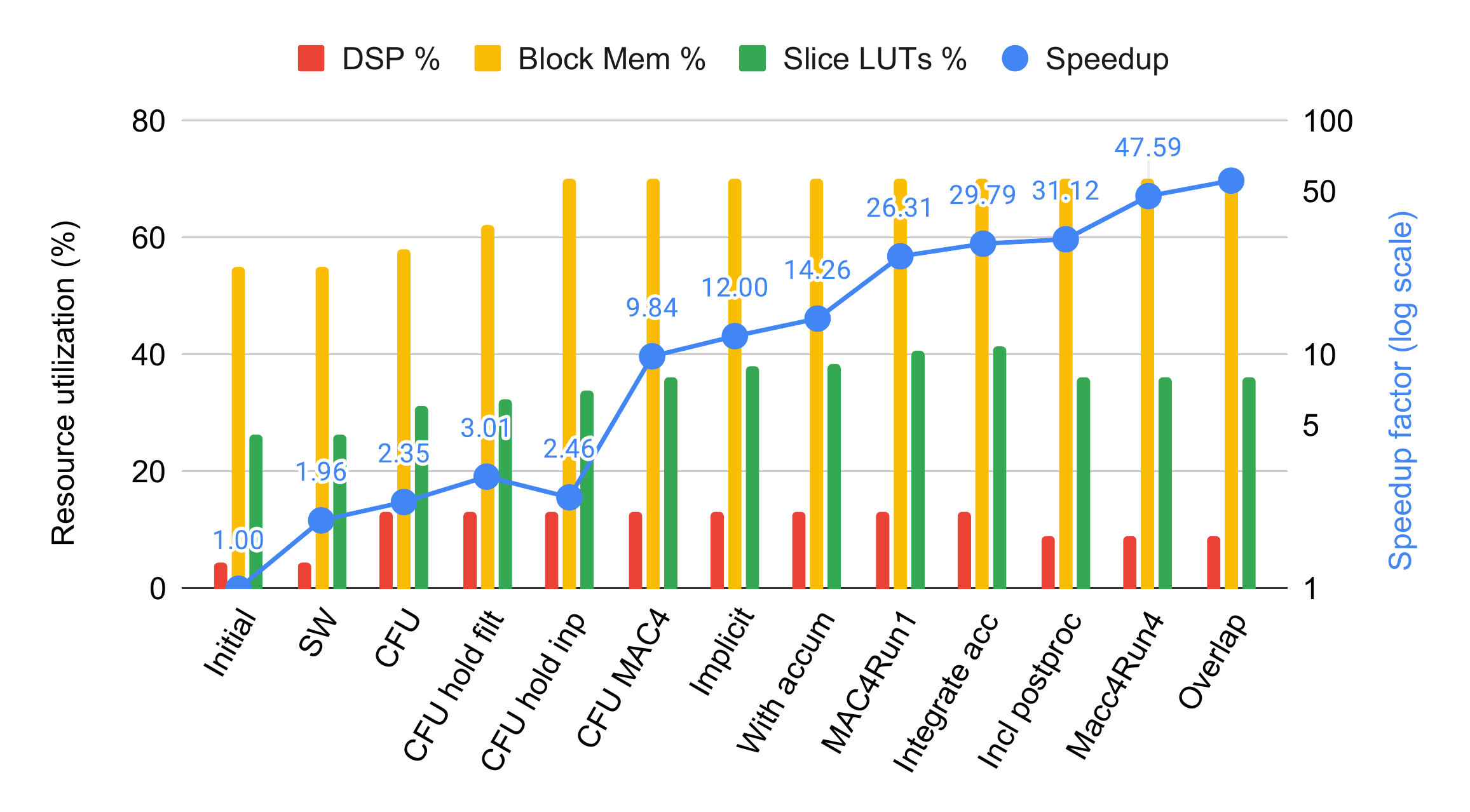}
  \caption{MobileNetV2— 1x1 \texttt{CONV\_2D} speedup.}
  \label{fig:mnv2_progression}
\end{figure}

\textbf{Profile:}
Profiling of MNV2 on the Arty board showed that the unaccelerated baseline application took about 900~M cycles. About 95\% of its execution time was spread across three different types of convolutions: 1x1 2D Convolution (63\%), Depthwise Convolution (22.5\%), and 3x3 2D Convolution (11\%).
Since most of the cycles were consumed by the \texttt{CONV\_2D} 1x1 layers, our primary focus was on accelerating that particular operator. In the computation for a 1x1 convolution, for each $x$, $y$ spatial coordinate, an input vector (the input tensor’s column at $x$, $y$ with the length determined by the number of input channels) is multiplied by a matrix to produce an output vector (the output tensor’s column at $x$, $y$ with length equal to the number of output channels). The matrix contains the filter weights, has size input channels $\times$ output channels, and is the same for every $x$, $y$. Post-processing (applying bias, activation function, and requantization) is applied to each output element. 

\textbf{Optimizations:}
Figure~\ref{fig:mnv2_progression} shows the speedup and resource usage as we stepped through each of the optimizations, iteratively. The optimization labels in Figure~\ref{fig:mnv2_progression} will be used for reference when describing them in the text. We began with software optimizations to minimize hardware changes until needed, and only then moved on to CFU hardware support. The CFU was implemented using \camera{Amaranth HDL.} 


\textbf{Software Optimizations and Specialization:}
We started by creating a new \texttt{CONV\_2D} kernel {\em specialized} for the 1x1 case, that is, when \texttt{filter\_width} and \texttt{filter\_height} are both equal to 1. A check in the general kernel branches to the specialized kernel when this is the case. Then in the specialized kernel, \texttt{filter\_width} and \texttt{filter\_height} can be assumed to be 1, and we can remove two levels of looping as well as replace other uses of those parameters with a constant 1. For example, a padding out-of-bounds check can be also be removed.
These specializations in addition to other optimizations such as loop unrolling reduced the execution time by 49\%, providing an almost 2$\times$ speedup (\textit{SW} in Fig.~\ref{fig:mnv2_progression}).

\begin{figure}[t!]
 \centering
  \includegraphics[trim=10 40 60 50, clip, width=.8\linewidth]{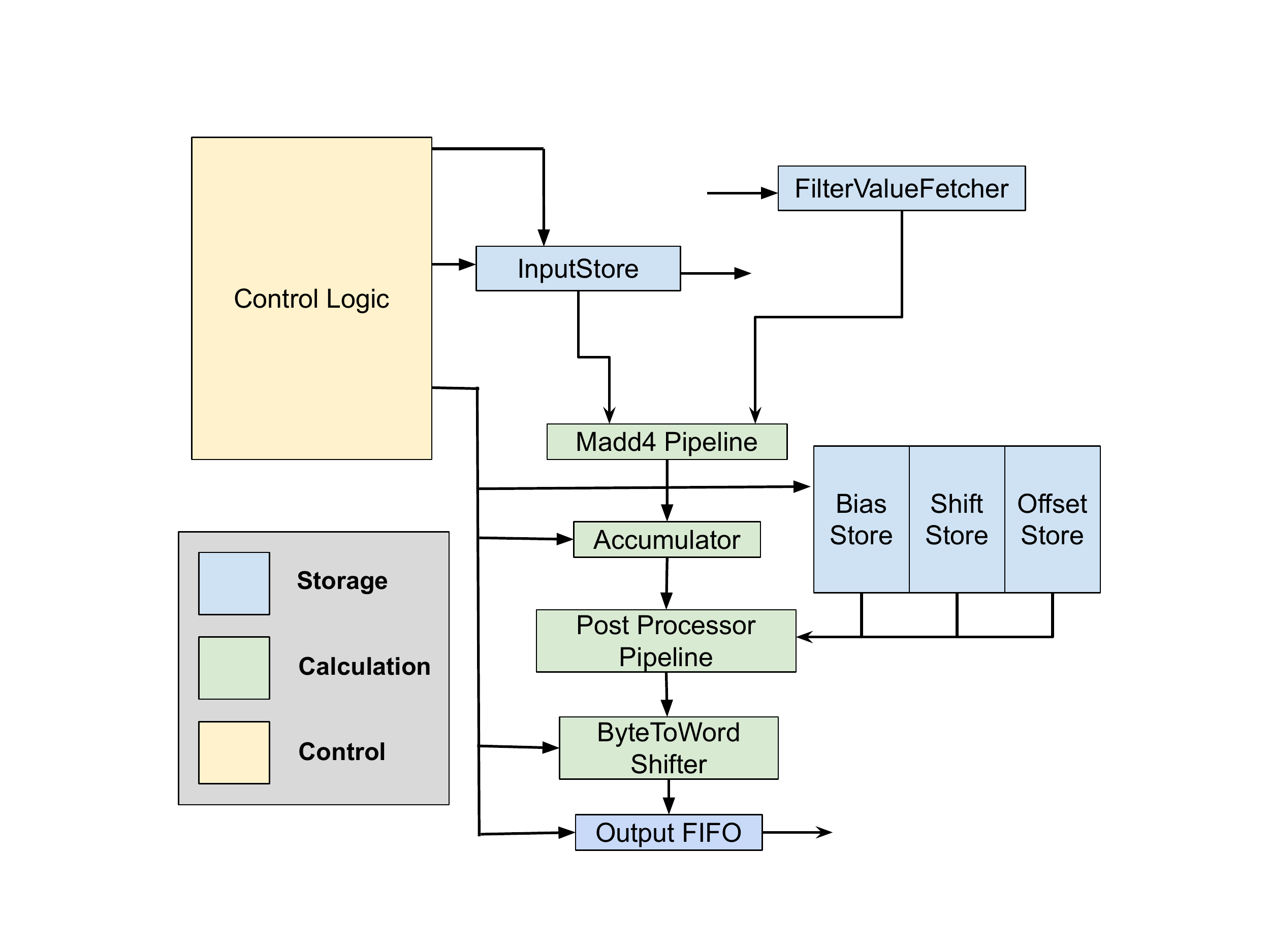}
 \caption{MNV2 CFU control logic and datapath design.}
 \label{fig:mnv2_design} 
\end{figure}

\textbf{CFU Optimizations:} Our first custom instruction accelerated the post-processing of each output element.
Figure~\ref{fig:mnv2_design} shows the CFU microarchitecture.
Per-output channel values for bias, multiplicand, and shift amount were stored in the CFU. This saved approximately 55 cycles per output, bringing the speedup to 2.3$\times$ (\textit{CFU}).

Next, moving filter values into the CFU (using the CFU as a small scratchpad memory) resulted in a small speed improvement---approximately 2 cycles per multiply-accumulate (MAC) (\textit{CFU hold filt}). However, moving input values into the CFU cost about 2 cycles per MAC. Since the CFU stores values by word instead of byte, the CPU must perform bit shifts and sign extensions to use values retrieved from the CFU. This canceled the speed up from reduced memory access and better data cache behavior (\textit{CFU hold inp)}.

However, the payoff came in the next step when we built a 4x4 multiply-accumulate (MAC4) instruction that operates on the packed filter and inputs retrieved from the CFU buffers. Even though the CPU is still involved in this data movement, the cumulative speedup reached 9.8$\times$ (\textit{CFU MAC4}). We then change the MAC4 instruction to pull input parameters directly from the previously constructed buffers and move the whole inner accumulation loop into the CFU. These changes got us down to less than one cycle per MAC, a cumulative speedup of 26$\times$ (\textit{MAC4Run1}). 
Finally, we connected the accumulation result directly to post-processing in the CFU without CPU intervention to reach 31.1$\times$ speedup (\textit{Incl postproc}).

Profiling the optimized version at this point showed that calculating and writing back 8-bit output channel values one at a time was not making efficient use of memory bandwidth, so we restructured the CFU to pack four of these outputs into a single 32-bit word for the CPU to write back to memory
(\textit{Macc4Run4}).
Finally, we pipelined the CFU to calculate while loading inputs to hide the input loading overhead  (\textit{Overlap}). We refer to this complete, large CFU design in later sections as \textit{CFU1}.

\textbf{Summary:} Overall, the MNV2-specialized design achieved a 55$\times$ speedup for 1x1 \texttt{CONV\_2D} over the original implementation.\footnote{For an {\em overall} speedup of this magnitude, we would also need to speed up the other significant operator types by a similar amount, which we have not yet implemented. Our overall speedup as a result for MNV2 was 3$\times$.} 
Throughout the iterative development process, we were never close to running out of any FPGA resources (as shown in Figure \ref{fig:mnv2_progression}). Resource usage peaked midway when the different processing steps were individually implemented on the CFU. As the processing became more integrated on the CFU, pathways for moving values back and forth to the CPU were removed, resulting in overall resource usage reduction.


\subsubsection{Keyword Spotting Acceleration on Fomu}
\label{sec:kws_on_fomu}

Keyword Spotting (KWS) is ubiquitous and an always-on  TinyML use case, making it a perfect candidate for acceleration. This example specifically demonstrates resource allocation optimization and trade-offs among the CPU, memory system, and CFU on a resource-constrained TinyML device. We show how we used CFU Playground's deploy$\rightarrow$profile$\rightarrow$optimize loop to accelerate the MLPerf~Tiny KWS quantized (int8) model by 75$\times$ with model-specific optimizations and a custom CFU.

\textbf{Target Objective:}
We started with a baseline that was  75$\times$ slower than CMSIS-NN hand optimized kernels for ARM Cortex-M CPUs\cite{lai2018cmsis}. The goal was to make the cycle count for our implementation comparable to such optimized kernels.

\textbf{Deploy:} We deployed the system to the tiny Fomu FPGA board, which is roughly the size of a penny and fits inside a USB slot. It combines an iCE40UP5k FPGA (with 5280 logic cells and 128~kB of on-chip RAM) with a 2~MB flash memory. 

\textbf{Profile:} The minimal VexRiscv configuration (without caches, hardware multiplication, branch prediction, or bypassing) did not fit on Fomu. To squeeze VexRiscv onto the FPGA we needed to remove features from the LiteX SoC (i.e., hardware timer and reset registers) and reallocate logic cell usage in the VexRiscv core by removing hardware error checking (e.g., for misaligned addresses). Iterative hardware multiplication was added for performance, but not division, which was handled by software emulation. These modifications were feasible due to the open-source nature of VexRiscv and LiteX. 
Detailed knowledge of the CPU was not required as changes were performed by simple configuration options.

Furthermore, the CFU Playground compiled binary image would not fit in 128kB, considering that much of this RAM is needed by TFLite Micro for working data. We modified the linker script to place the code (\texttt{.text} section) and read-only data (\texttt{.rodata} section— mostly weights from the ML models) into flash memory. The KWS application finally fit but took 2.5 minutes to run, which we eventually optimized to run in 2 seconds.

\begin{figure}[t]
\vspace{-.5em}
  \centering
  \includegraphics[width=\linewidth]{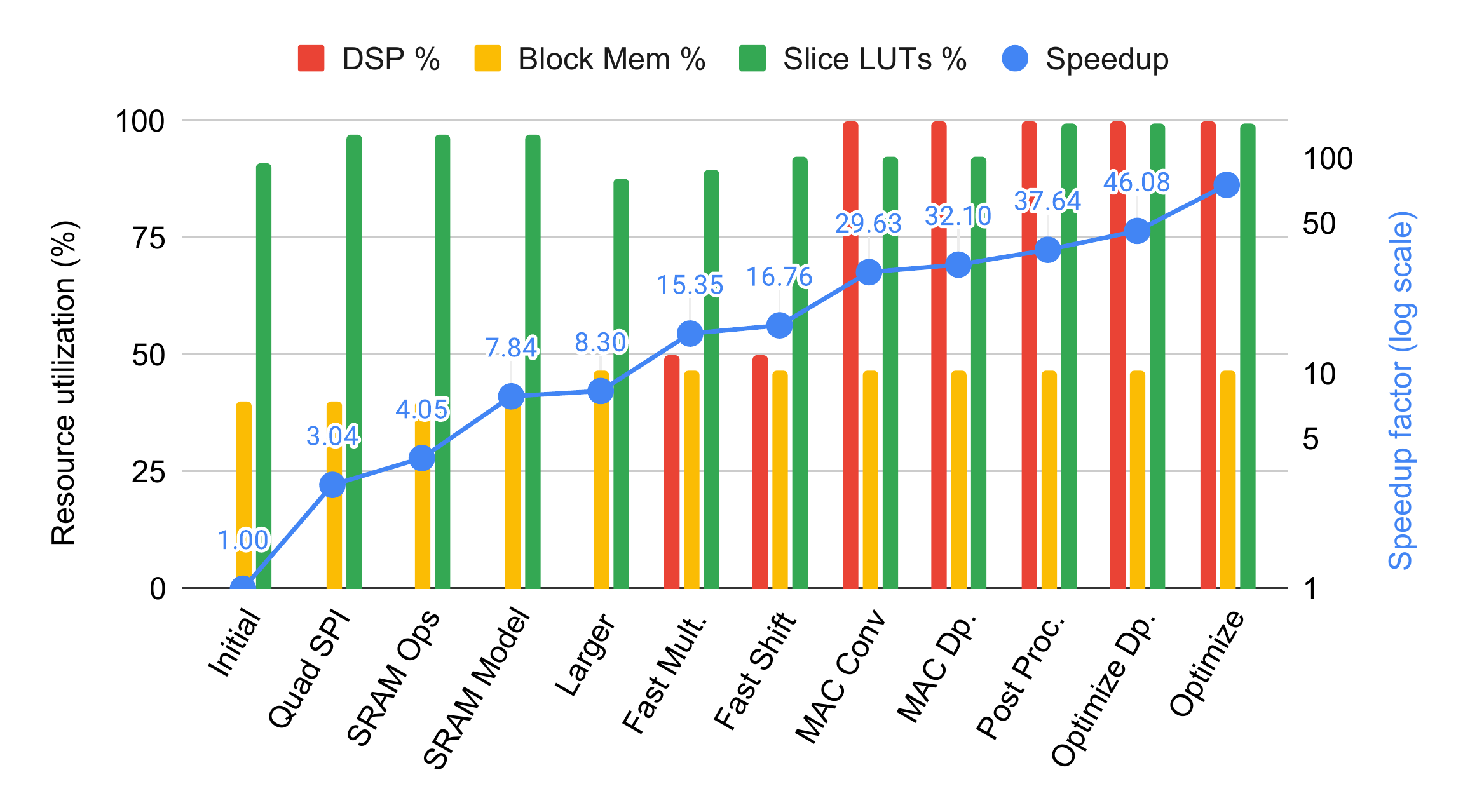}
  \caption{Speedup and resource usage for the KWS use case.}
  \label{fig:kws-speedup} 
\end{figure}

\textbf{Memory System Optimizations:}
Profiling showed that
the flash ROM accesses were slower than they should be. This pointed to some potential improvements in the SPI flash interface.
First, we upgraded the ROM interface from a Serial Peripheral Interface (SPI) to a Quad SPI, substantially improving the ROM read bandwidth,
giving a 3.04$\times$ speedup over the baseline (\textit{QuadSPI} in Fig.~\ref{fig:kws-speedup}).

Next, we moved critical sections from the ROM to the SRAM. Since SRAM capacity was severely limited (128~kB), we selected the primary bottlenecks, which were determined via profiling. These included the code for \texttt{CONV\_2D}, \texttt{Depthwise\_CONV\_2D}, and model weights. This optimization led to a total speedup of 7.84$\times$ (\textit{SRAM Ops and Model}).

We then removed unnecessary control/status registers and SoC features intended for debugging to make space for a larger I-Cache that decreased the average instruction fetch time. This led to a performance speedup of 8.3$\times$ (\textit{Larger I-Cache}).

\textbf{CFU Optimizations:}
Owing to the previous resource savings, we now had room to add single-cycle multiplication, rather than use our iterative $\sim$30 cycle multiplication in our CPU. This used four of Fomu's eight DSP tiles and increased performance by 15.35$\times$ (\textit{Fast Mult}). The savings also created room to add a 4-way parallel MAC CFU to address the convolution bottleneck (\textit{MAC Conv)}, which required the remaining four DSP tiles. Depthwise convolution was the second runtime contributor. It has a different memory access pattern and thus could use the 4-way multiply-accumulate that we built for the convolution. Ideally, we could build separate CFU gateware for depthwise convolution, but there were no remaining resources to extend the CFU this way. Instead, we utilized a single lane of the 4-way multiply-accumulate CFU in the depthwise convolution to achieve a cumulative speedup of 32.10$\times$.


After implementing the SIMD MAC operation, there were logic cells remaining (although no DSP tiles were left). With these logic cells, we added extra functionality to the CFU to perform accumulator post-processing (consisting of saturating multiplication, rounding division, and output clamping) 14$\times$ faster deep inside the convolution and depthwise convolution operations. This change led to a 37.64$\times$ speedup (\textit{Post Proc}). 



\textbf{Software Optimizations:} On the software front, we specialized the operators. The convolution operators in TensorFlow Lite are quite generalized. Therefore, we informed the compiler about constants and invariants (our \texttt{filter\_width} is always 3, our \texttt{depth\_multiplier} is always 1, etc.). This lets the compiler generate more efficient assembly, decreasing the number of branches (which are expensive as our MCU-class CPU does not have room for a predictor here). 

\textbf{Summary:} KWS performance improved by a factor of 75$\times$ from the baseline. The time for one inference reduced from 2.5 minutes to under 2 seconds on the FPGA. An ASIC implementation of the design would be an order of magnitude faster, but that is outside the scope of this work. Only 3$\times$ of the speedup was directly attributed to the small CFU. We refer to this CFU design in the next section as \textit{CFU2}. The other 25$\times$ was derived from optimizing the CPU, software, memory accesses, and system interfaces \& drivers. The final optimized Fomu KWS results, if normalized for the differing clock rates, are roughly comparable to the MLPerf Tiny results~\cite{banbury2021mlperf} for the much more complex Cortex-M4 with hand-optimized CMSIS-NN kernels utilizing the M4 SIMD instructions. We stopped once we reached this state of the art solution but could have kept making improvements using the tool.
This illustrates the usefulness of CFU Playground’s iterative design loop running on a live system, facilitating hardware-software co-design.




\subsection{Automated Design Space Exploration of CPU vs. CFU}
\label{sec:dseengine}

\begin{figure}[t]
\vspace{-.5em}
  \centering
  \includegraphics[width=\linewidth]{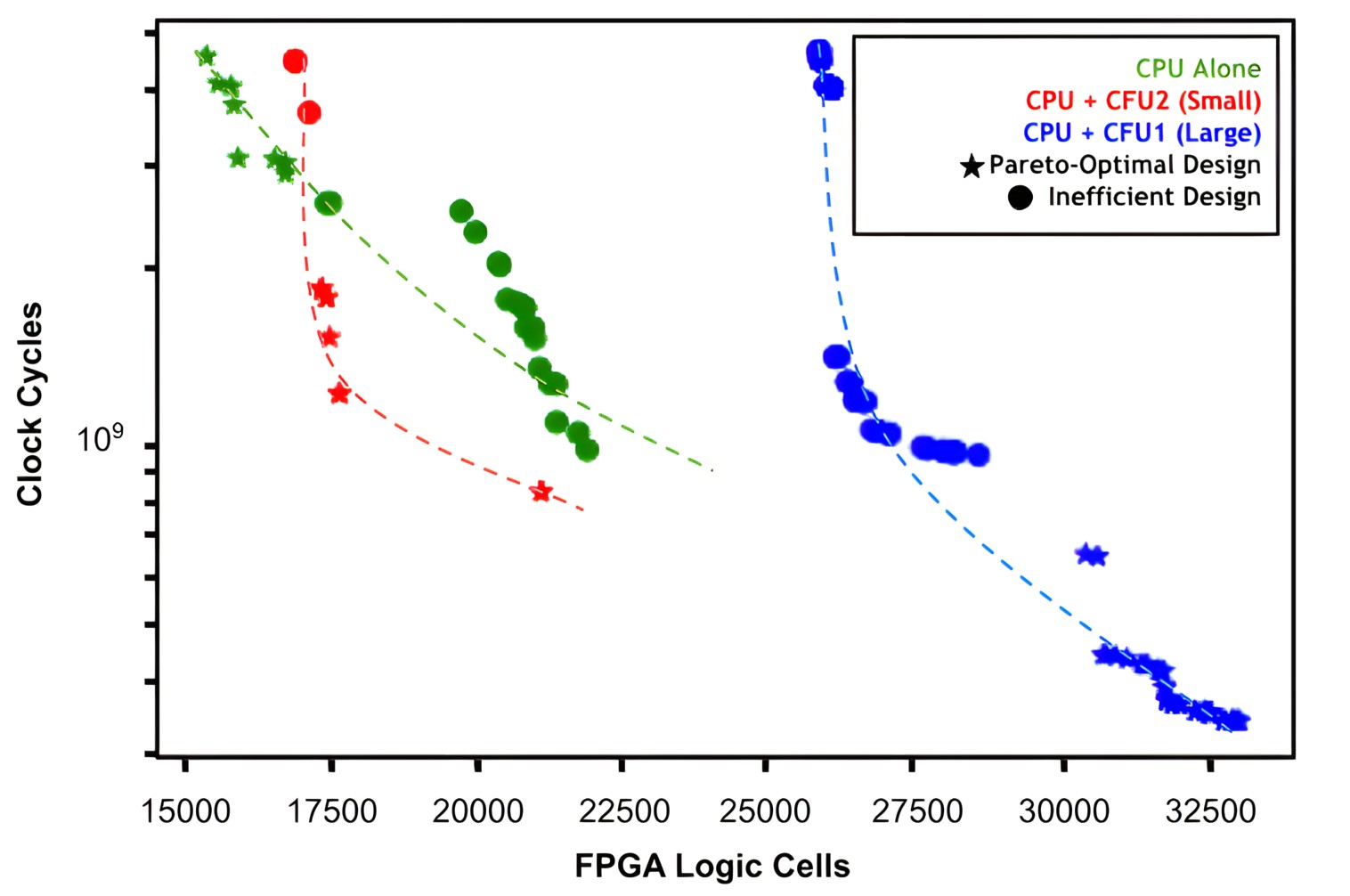}
   \caption{\new{Pareto curves formed by Vizier's design space exploration. X-axis represents resource consumption on the FPGA board via logic cells (less is better). Y-axis represents workload latency via cycle count (less is better). Overall Pareto-optimal points in design space are starred.}}
  \label{fig:pareto-curve} 
\vspace{-.1em}
\end{figure}

 

After building the CFUs mentioned in the two use cases in Section~\ref{sec:case_studies} (i.e., \textit{CFU1} and \textit{CFU2}), we used the DSE capabilities built into CFU Playground to explore the design space of different CPU + CFU configurations for the MNV2 workload specifically. The design space included approximately $93,000$ different design points when considering all of the soft CPU's configurable architectural parameters combined with the CFUs.
Figure~\ref{fig:pareto-curve} shows the DSE performed by Vizier. 

To scale the number of different design points Vizier could test, we ran the experiments in the cloud using simulation. 
There are three different Pareto curves in Figure~\ref{fig:pareto-curve}. The green curve is for the CPU alone, the blue is for the CPU \camera{coupled with the large, complex} CFU1, and the red is for the CPU \camera{with the much smaller and simpler} CFU2. The starred points highlight the overall Pareto-optimal points. These results demonstrate how CFU designs can create a richer design space, leading to more optimal configurations. For example, on the plot we can see there exist design points on the red curve (\textit{CPU + CFU2}) that consume \textit{fewer} resources but are \textit{more} performant than some of the larger design points on the green curve (\textit{CPU Alone}).

\section{\new{Prior} Work}
\label{sec:related}

\newcommand{\greencheckmark}{{\color{green}\cmark}}
\newcommand{\redxmark}{{\color{red}\xmark}}
\begin{table*}[t!]
\centering
\resizebox{\linewidth}{!}{
\begin{tabular}{|c|c|c|c|c|c|c|c|c|c|c|}
 \hline
 & 
 \shortstack{Open \\Source} & 
 \shortstack{Full \\Stack} & 
 \shortstack{Full \\SoC} &
 \shortstack{Tightly Coupled/ \\Specialized ISA} & 
 \shortstack{Fine-Grained \\\camera{Accelerated} ML Ops} & 
 \shortstack{Hardware \& Engineer \\In-The-Loop} &  
 \shortstack{Stock \\Compiler} & 
  \shortstack{\camera{Automated CPU$\leftrightarrow$Accelerator} \\\camera{Design Space Exploration}} & 
 \shortstack{TinyML \\Focus}
 \\ [0.5ex] 
 \hline
 
\hline\hline
 \textbf{CFU Playground} & \greencheckmark & \greencheckmark & \greencheckmark & \greencheckmark & \greencheckmark & \greencheckmark & \greencheckmark & \greencheckmark  & \greencheckmark\\ 
\hline\hline

\camera{Chipyard}\cite{amid2020chipyard} & \greencheckmark & \greencheckmark & \greencheckmark & \greencheckmark & \greencheckmark & \greencheckmark &\greencheckmark  & \redxmark & \redxmark\\ 
\hline

\hline
\camera{Centrifuge}\cite{huang2019centrifuge} & \greencheckmark & \greencheckmark & \greencheckmark & \greencheckmark & \greencheckmark & \greencheckmark &\redxmark  & \redxmark & \redxmark\\ 
\hline

\hline
 \camera{Embedded Scalable Platform}\cite{mantovani2020agile} & \greencheckmark & \greencheckmark & \greencheckmark & \redxmark & \redxmark & \greencheckmark &\greencheckmark  & \redxmark & \redxmark\\ 
\hline

\hline
 Gemmini\cite{genc2021gemmini} & \greencheckmark & \greencheckmark & \greencheckmark & \redxmark & \redxmark & \redxmark & \greencheckmark  & \redxmark & \redxmark\\
\hline

\hline
 hls4ml\cite{DBLP:journals/corr/abs-2103-05579} & 
 \greencheckmark & \redxmark & \greencheckmark & \redxmark & \redxmark &\greencheckmark & \redxmark & \redxmark & \greencheckmark\\ 
\hline

\hline
 Deepburning\cite{wang2016deepburning}& \greencheckmark & \greencheckmark & \greencheckmark & \redxmark & \redxmark & \redxmark &\redxmark  & \redxmark & \redxmark \\ 
\hline

\hline
 DNN-Weaver\cite{sharma2016dnnweaver}& \greencheckmark & \greencheckmark & \redxmark & \greencheckmark & \redxmark & \redxmark & \redxmark  & \redxmark & \redxmark \\
\hline

\hline
  DNN-Builder\cite{8587697} & \greencheckmark & \greencheckmark & \redxmark & \redxmark  & \redxmark & \redxmark & \redxmark  & \redxmark & \redxmark \\
\hline

\hline
 FINN\cite{umuroglu2017finn} & \greencheckmark & \redxmark & \redxmark & \redxmark & \redxmark & \redxmark &\redxmark  & \redxmark & \redxmark\\ 
\hline





\end{tabular}
}
\vspace{12pt}
\caption{Comparison of CFU Playground with open-source toolchains \camera{supporting custom hardware design for ML workloads}.
CFU Playground focuses on \textit{open-source} development across the \textit{full} \textit{system} stack, while providing varying levels of flexibility for hardware and software (co-)design.}
\label{table:FPGAToolflowComparison}
\end{table*}


\new{In recent years, FPGA-based ML accelerators have gained momentum in the race to speed up ML tasks because of their efficiency compared to GPUs, lighter investment than ASICs, and overall flexibility provided by their ability to be reconfigured \cite{10.1145/3020078.3021736}. As a result, many tools have been built to support FPGA design for ML. However, doing a direct head-to-head quantitative comparison against other tools is challenging, though, as we believe there is no ``one size fits all'' solution for designing custom ML \camera{hardware} accelerators. The best choice of workflow is task-dependent and up to the developer. Therefore, we give a qualitative comparison of CFU Playground against prior work along different dimensions. CFU Playground adds a dimension of flexibility on top of prior art, especially useful in resource-constrained embedded systems where discrete accelerators are not always feasible.}

There have been other tools and workflows for accelerating \camera{ML with custom hardware}. 
Table ~\ref{table:FPGAToolflowComparison} compares the features supported by some of the existing frameworks compared to CFU Playground.
More comprehensive and extensive surveys of existing flows for FPGAs can be found in the related work~\cite{shawahna2018fpga, venieris2018toolflows, guo2019dl}.
Specialized ISAs have been explored for accelerating DNNs using application-specific instruction-set processors (ASIPs)~\cite{yang2021flexacc, liu2016cambricon}. 
\new{These approaches developed fixed, domain-specific ISAs for more diverse and generalized DNN acceleration, in contrast to our extreme model-specific customization, useful for embedded systems which are often task-specific and less aimed for general-purpose computation.}
Also, the tools used for ASIP accelerators are commercial~\cite{synopsys, cadence} and require costly ramp-up. 
\new{These tools require a lot more overhead for users to begin developing and deploying in comparison to CFU Playground's working, out-of-the-box solution that includes a full ML inference stack and SoC already pieced together and running. CFU Playground is a full-stack end-to-end framework with designs that can be readily adopted and reproduced.}

\new{Furthermore, it is common to not tightly couple the accelerator with the CPU, and most workflows follow this design pattern~\cite{umuroglu2017finn,DBLP:journals/corr/abs-2103-05579,8587697,wang2016deepburning,genc2021gemmini, mantovani2020agile}. Due to this decoupling, many flows design the accelerators in isolation \textit{before} integrating with the rest of the system and stack. This can cause a failure post-design due to lack of accounting for effects that arise elsewhere in the computing stack or at the system level, such as the cost of offloading, scheduling, etc.}
Designing the accelerator separate from the system does not allow for co-design with the processor to explore trade-offs with customizing the CPU. 
\camera{Centrifuge~\cite{huang2019centrifuge}, for example, does not include the CPU in the exposed hardware design space as it is in CFU Playground.} 

\camera{Chipyard~\cite{amid2020chipyard} does enable end-to-end integration of tightly coupled accelerators but does not explore design space tradeoffs between CPU and accelerator or provide automatic parameter exploration as our solution does.}
Gemmini \camera{is one work to briefly} mention design trade-offs between accelerator and CPU but uses an architectural template to generate \camera{systolic array} accelerators together with a software stack and integrated SoC to capture system-level effects ~\cite{genc2021gemmini}. 

However, similar to a majority of the accelerator workflows, \camera{Gemmini's template} is designed to optimize and run \camera{general matrix multiplication kernels}.
hls4ml~\cite{DBLP:journals/corr/abs-2103-05579} is a complete tool for deploying different models to FPGAs, but the accelerators \camera{are also designed to speed up large kernels or the entire model} rather than individually selected ML operations. CFUs provide finer-granularity \camera{than these approaches} and can be designed to accelerate \camera{model-}specific operations \textit{within} ML kernels incrementally as well as entire kernels, while also supporting the rest of the system stack that would normally be deployed on-device. Finally, most other design flows mentioned in Table~\ref{table:FPGAToolflowComparison} auto-generate hardware implementations of the accelerators using fixed templates and specialized compilers. These are limiting and less flexible than CFU Playground's largely engineer in-the-loop approach and use of RISC-V. Moreover many flows are often not model-agnostic, but rather limited to a specific class of ML models such as FINN~\cite{umuroglu2017finn}.

\section{Conclusion and Future Work}


\new{TinyML requires bespoke architectures that rely on hardware-software co-design. To that end, CFU Playground provides users an open-source full-stack framework for experimental and novel model-specific acceleration. We expose developers to a fast and iterative design and evaluation flow to obtain significant speedups by exploring the design space between the CPU and a tightly-coupled CFU. This is specifically helpful for heavily resource-constrained platforms that cannot afford discrete accelerators for power and area constraints.}


\new{CFU Playground has been adopted and supported by the industry and it will continue to stay relevant. We will continue to evolve CFU Playground with an open-source ASIC flow, additional model architecture support, direct memory access for the tightly-coupled CFU engine, and more open-source \camera{RISC-V} CPUs \camera{(soft and hardened)} with the CFU interface. \camera{Investigation of high-level synthesis (HLS) pragmas for identifying where execution hotspots exist to automatically generate CFUs and insert calls to custom instructions could provide further automation.} Future work will also involve studying the optimization space for power and energy efficiency.}


\bibliographystyle{IEEEtran}
\bibliography{refs}

\end{document}